\RequirePackage{snapshot}

\documentclass{article}

\usepackage[nonatbib,preprint]{nips_2018}

\usepackage[utf8]{inputenc} \usepackage[T1]{fontenc}    \usepackage{hyperref}       \hypersetup {
  colorlinks = True,
}
\usepackage{url}            \usepackage{booktabs}       \usepackage{amsfonts}       \usepackage{nicefrac}       \usepackage{microtype}      

\usepackage{graphicx}
\usepackage{amsmath}
\usepackage{amssymb}
\usepackage{amsthm}
\usepackage[font=small]{caption}
\usepackage{subcaption}

\usepackage{tablefootnote}
\usepackage{adjustbox}
\usepackage{textcomp}  \usepackage{xcolor,colortbl}  \usepackage[para]{footmisc}  

\newcommand{\ignore}[1]{}

\usepackage{color}

\renewcommand{\footnotesize}{\scriptsize}

\definecolor{Gray}{gray}{0.9}
\newcolumntype{g}{>{\columncolor{Gray}}c}  \newcolumntype{H}{>{\setbox0=\hbox\bgroup}c<{\egroup}@{}}  \definecolor{GrayLine}{gray}{0.7}

\usepackage[
  capbesideposition={right,center},
  capbesidesep=quad,
]{floatrow}

\newcommand{\testap}{21.91}
\newcommand{\valap}{22.8}

\usepackage{cite}

\title{A Better Baseline for AVA}

\author{Rohit Girdhar$^{\dag\ddag}$ \quad   Jo{\~ a}o Carreira$^{\dag}$ \quad Carl Doersch$^{\dag}$ \quad Andrew Zisserman$^{\dag *}$
\\    $^{\dag}$DeepMind \quad $^{\ddag}$Carnegie Mellon University \quad $^{*}$University of Oxford}

\begin{document}
  
\maketitle

\begin{abstract}

  We introduce a simple baseline for action localization on the AVA dataset.
  The model builds upon the Faster R-CNN bounding box detection framework,
  adapted to operate on pure spatiotemporal features -- in our case produced exclusively by an I3D model pretrained on Kinetics. 
  This model obtains 21.9\%   average AP on the validation set of AVA v2.1,
  up from 14.5\% for the best RGB spatiotemporal model used in the original AVA paper
  (which was pretrained on Kinetics and ImageNet),
  and up from 11.3\% of the publicly available baseline using a ResNet-101 image feature extractor,
  that was pretrained on ImageNet.
  Our final model obtains 22.8\%/21.9\% mAP on the val/test sets and { outperforms all submissions} to the AVA challenge at CVPR 2018.

\end{abstract}

\section{Introduction}

  Despite considerable advances in the ability to estimate position and pose for people and objects, the computer vision community lacks models that can describe what people are doing at even short-time scales. 
  This has been highlighted by new datasets such as Charades \cite{sigurdsson2016hollywood} and AVA \cite{gu2017ava}, where the goal is to recognize the set of actions people are doing in each frame of example videos -- e.g. one person may be standing and talking while holding an object in one moment, then it puts the object back and sits down on a chair. 
  The winning system of the Charades challenge 2017 obtained just around 21\% accuracy on this per-frame classification task. 
  On AVA the task is even harder as there may be multiple people and the task additionally requires localizing people and describing their actions individually -- a strong baseline gets just under 15\% on this task \cite{gu2017ava}. 
  The top approaches in both cases used I3D models trained on ImageNet \cite{russakovsky2015imagenet} and the Kinetics-400 dataset \cite{kay2017kinetics}.

  In this work, we focus on diagnosing and improving that system by carefully examining the various design decisions that go into building a video action localization model. Specifically, we find data augmentation, class agnostic bounding box regression and pre-training lead to strong performance gains on AVA. Our resulting model outperforms all previous approaches, including all submissions to the AVA challenge at CVPR 2018. This includes various highly sophisticated solutions involving multiple input modalities like optical flow and audio, as well as ensembles of multiple network architectures.

\section{Model and Approach}

  Our model is inspired by I3D~\cite{carreira2017quo} and Faster R-CNN~\cite{ren2015faster}, similar to~\cite{girdhar2018detecttrack,hou2017tube}.
  We start from labeled frames in the AVA dataset, and extract a short video clip, typically 64 frames, around that keyframe. 
  We pass this clip through I3D blocks up to {\tt Mixed\_4f}, which are pre-trained on the Kinetics dataset for action classification.
  The feature map is then sliced to get the representation corresponding to the center frame (the keyframe where the action labels are defined). 
  This is passed through the standard region proposal network (RPN)~\cite{ren2015faster} to extract box proposals for persons in the image. 
  We keep the top 300 region proposals for the next step: extracting features for each region that feed into a classifier.

  Since the RPN-detected regions corresponding to just the center frame are 2D, we extend them in time by replicating them to the match the temporal dimension of the intermediate feature map, following the procedure for the original AVA algorithm~\cite{gu2017ava}. 
  We then extract an intermediate feature map for each proposal using the RoIPool operation, applied independently at each time step, and concatenated in time dimension to get a 4-D region feature map for each region.
  This feature map is then passed through the last two blocks of the I3D model (up to {\tt Mixed\_5c}, and classified into each of the 80 action classes.
  The box classification is treated as a non-exclusive problem, so probabilities are obtained through an independent sigmoid for each class.
  We also apply bounding box regression to each selected box following Faster R-CNN~\cite{ren2015faster}, except that our regression is independent of category (since the bounding box should capture the person regardless of the action).
  Finally we post-process the predictions from the network using non-maximal suppression (NMS), which is applied independently for each class. 
  We keep the top-scoring 300 class-specific boxes (note that the same box may be repeated with multiple different classes in this final list) and drop the rest.

\begin{figure}[t]
    \centering
    \includegraphics[width=\linewidth]{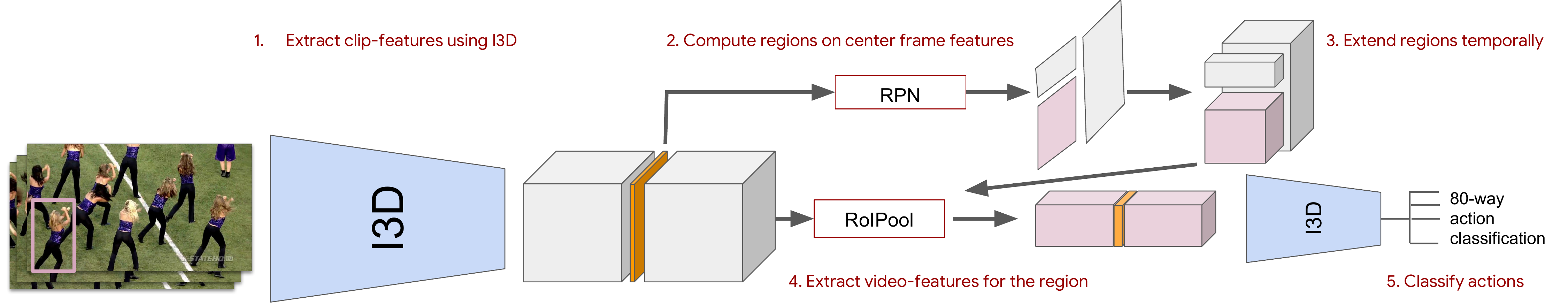}
    \caption{{\bf Network architecture.} We build upon I3D and Faster R-CNN architectures. A video clip is passed through the first few blocks of I3D to get a video representation. The center frame representation then is used to predict potential `person' regions using a region proposal network (RPN). The proposals are extended in time by replicating, and used to extract a feature map for the region using RoIPool. The feature map is then classified into the different actions using two I3D blocks.}
    \label{fig:nwarch}
\end{figure}

\section{Experiments}

\begin{table}
\floatbox[\capbeside]{table}
{\caption{{\bf Validation set results.} Here we compare our model to some of the previously proposed methods in literature. Our simple model outperforms all previous approaches by a significant margin.}\label{the_table}}{\begin{center}
\begin{tabular}{cc}
\toprule
Method & Validation mAP \\
\midrule
ResNet-based model~\cite{ava_baseline} & 11.3 \\
RGB only~\cite{gu2017ava} & 14.5 \\
RGB + Flow~\cite{gu2017ava} & 15.6 \\
Ours & 21.9 \\
Ours + JFT & {\bf \valap{}} \\
\bottomrule
\end{tabular}
\end{center}}
\end{table}

We trained the model on the training set using a synchronized distributed setting with 11 V100 GPUs. 
  We used batches of 3 videos with 64 frames each, and augmented the data with left-right flipping and spatial cropping. 
  We trained the model for 500k steps using SGD with momentum and cosine learning rate annealing. 
  Before submitting to the challenge evaluation server, we finetuned the model further on the union of the train and validation sets. 
  We tried both freezing batch norm layers and finetuning them with little difference in performance.
  For the experiments of training the model from scratch, we train the batch norm layers as well. Since that leads to higher memory usage, we use a batch size of 2 and train over 32 GPUs (for an effectively similar batch size).

  Results of our model on the validation set are compared with results from the models in the AVA paper in Table~\ref{the_table}. 
  The RGB-only baseline~\cite{gu2017ava} used a similar I3D feature extractor similar to ours, but pretrained on ImageNet then Kinetics-400, whereas our model was just pretrained on Kinetics-400 or the larger Kinetics-600. 
  This baseline differs from ours in a few other ways: 
  1) it used a ResNet-50 for computing proposals and I3D for computing features for the classification stage, whereas we only use the same I3D features for both; 
  2) our model preserves the spatiotemporal nature of the I3D features all the way to the final classification layer, whereas theirs performs global average pooling in time of the I3D features right after ROI-pooling; 
  3) we opted for action-independent bounding box regression, whereas theirs learns 80 different regressors, one for each class.
  The RGB+Flow baseline is similar but also uses flow inputs and a Flow-I3D model, also pretrained on Kinetics-400. 
  The ResNet-101 baseline corresponds to a traditional Faster-RCNN object detector system applied to human action classes instead of objects, using just a single frame as input to the model.

  Our model achieves a significant improvement of nearly 40\% over the best baseline (RGB+Flow), while using just RGB and just one pretrained model instead 3 separate ones. 
  The results suggest that simplicity, coupled with a large pre-training dataset for action recognition, are helpful for action detection.
  This is reasonable considering that many AVA categories have very few examples, and so over-fitting is a serious problem.

\begin{figure}[t]
    \centering
    \includegraphics[width=\linewidth]{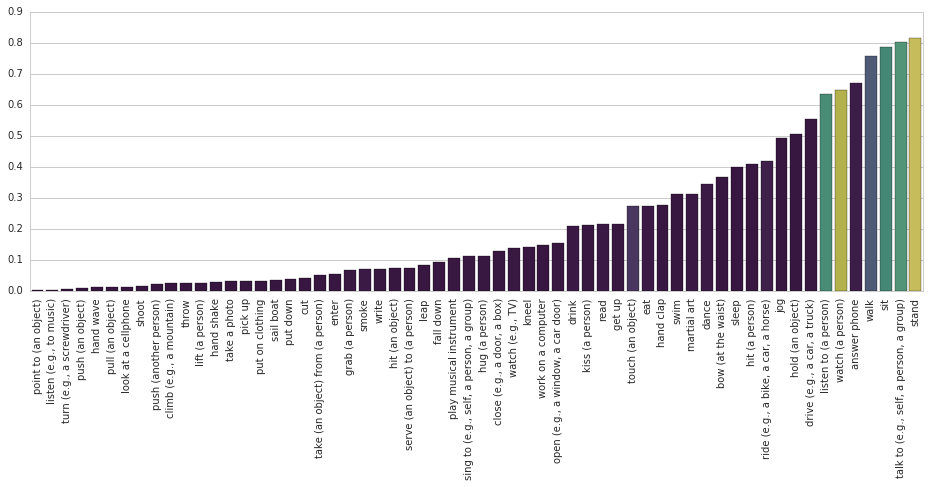}\hfill
    \caption{{\bf Per-class performance.} Performance of our model for each of the 60 action class evaluated in AVA. We color code the performance bars with the relative size of the class in the dataset (lighter colors are more common). As evident, there is some correlation of amount of data with the highest performing classes.}
    \label{fig:per-class}
\end{figure}
  
\begin{figure}[t]
    \centering
    \includegraphics[width=0.32\linewidth]{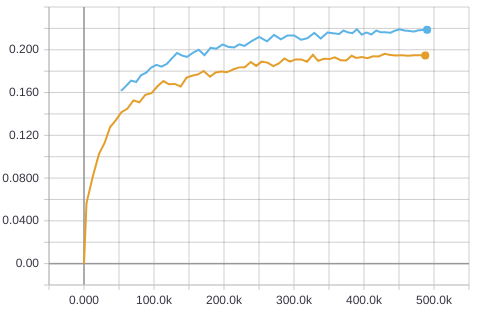}\hfill
    \includegraphics[width=0.32\linewidth]{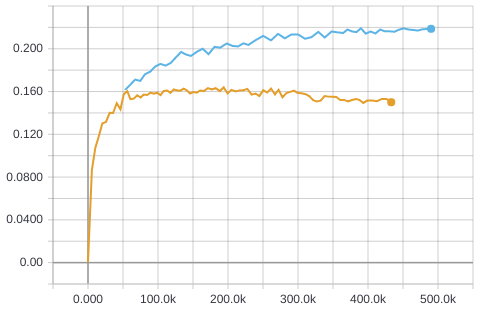}
    \includegraphics[width=0.32\linewidth]{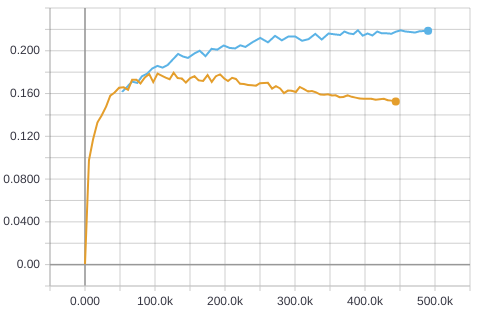}\hfill
    \caption{{\bf Ablations.} Validation performance curves of our model (blue) compared to baselines (yellow) over training iterations. Baselines here are the exact same model, with exactly one thing removed (in order): (i) Kinetics initialization; (ii) Single-class bounding box regression and (iii) Data augmentation. These three parameters were some of the important design decisions of our model.}
    \label{fig:ablations}
\end{figure}

We also formally ablate some of the important design decisions of our model in Figure~\ref{fig:ablations}. First, we evaluate the effect of initialization by comparing the scratch trained model with the Kinetics initialized model. We find about 2\% improvement on pretraining. Next we evaluate the importance of class agnostic bounding box regression compared to class-specific, and observe an almost 4\% gap in performance. This makes sense as our object is always a human, and it is a good idea to share the parameters for localizing a human across classes, as some of the smaller classes may not have enough data to learn an effective representation. Next we compare our model with and without data augmentation; in our case random flips and crops. This gives almost 5\% improvement, again signifying the importance of maximizing the amount of training data we can use.

\vspace{2mm}
{\noindent \bf Scene context:} To further incorporate context for recognizing actions, we experimented with adding full-image features for the key-frame when classifying each box in the video clip. We use the last layer features from a ResNet-101 pre-trained on the JFT dataset~\cite{sun2017revisiting}. We found that concatenating the 512D \texttt{global\_pool} features with the each box's features before classification gave a 0.9\% improvement on the val set, as shown in Table~\ref{the_table}. Hence we incorporate this in our final model.
Finally, we show the per-class performance of our model in Figure~\ref{fig:per-class}. We show our final test performance and comparison with other submissions in Table~\ref{tab:test}.

\begin{table}[t]
\setlength{\tabcolsep}{6pt}
\caption{{\bf Test set results.} Here we compare our method to the top submissions in the CVPR challenge. Again, our simple model outperforms all previous submissions, including ones involving multiple input modalities and network ensembles. The model abbreviations used here refer to the following. I3D: Inflated 3D convolutions~\cite{carreira2017quo}, FRCNN: Faster R-CNN~\cite{ren2015faster}, NL: Non-local networks~\cite{wang2017non}, P3D: Pseudo-3D convolutions~\cite{qiu2017learning}, C2D~\cite{tran2018closer}, C3D~\cite{tran2018closer}, TSN: Temporal Segment Networks~\cite{wang2016temporal} and FPN: Feature Pyramid Networks~\cite{lin2017feature}. Some of the submissions also attempted to use other modalities like audio, but got lower performance. Here we compare with their best reported numbers.}\label{tab:test}
\begin{center}
\resizebox{\textwidth}{!}{\begin{tabular}{cccc}
\toprule
Method & Modalities & Architecture & Test mAP \\
\midrule
Ours + JFT & RGB only & I3D + FRCNN & {\bf \testap{}} \\
\begin{tabular}{@{}c@{}} Ours (challenge submission) \end{tabular} & RGB only & I3D + FRCNN & 21.03 \\
\begin{tabular}{@{}c@{}} Tsinghua/Megvii (challenge winner) \end{tabular}
  & RGB + Flow &
  \begin{tabular}{@{}c@{}} I3D + FRCNN + NL + TSN + \\ C2D + P3D + C3D + FPN  \end{tabular}
       & 21.08 \\
  YH Technologies~\cite{yh_ava_submit} & RGB + Flow & P3D + FRCNN & 19.60 \\
  Fudan University & - & - & 17.16 \\
\bottomrule
\end{tabular}
}
\end{center}
\end{table}

\section{Conclusion}

  We have presented an action localization model that aims to densely classify the actions of multiple people in video using the Faster R-CNN framework with spatiotemporal features from an I3D model pretrained on the Kinetics dataset. 
  We show a large improvement over the state-of-the-art on the AVA dataset, but at \testap{}\% AP, performance is still far from what would be practical for applications.
  More work remains to be done to understand the problems in the model and dataset, such as handling classes with very small number of training examples.
  In the meanwhile, continuing to grow datasets such as Kinetics should help.

\paragraph{Acknowledgments:} Many thanks to the AVA team for creating and sharing their dataset, models and code.

{\small
\bibliographystyle{abbrv}  \bibliography{biblioLong,egbib}

\begin{thebibliography}{10}

\bibitem{ava_baseline}
{AVA Team}.
\newblock Ava v2.1 faster rcnn resnet-101 baseline.
\newblock \url{https://research.google.com/ava/download.html}.
\newblock Accessed: 2018-06-10.

\bibitem{carreira2017quo}
J.~Carreira and A.~Zisserman.
\newblock Quo vadis, action recognition? a new model and the kinetics dataset.
\newblock In {\em Proceedings of the IEEE Conference on Computer Vision and
  Pattern Recognition (CVPR)}, 2017.

\bibitem{girdhar2018detecttrack}
R.~Girdhar, G.~Gkioxari, L.~Torresani, M.~Paluri, and D.~Tran.
\newblock {Detect-and-Track: Efficient Pose Estimation in Videos}.
\newblock In {\em Proceedings of the IEEE Conference on Computer Vision and
  Pattern Recognition (CVPR)}, 2018.

\bibitem{gu2017ava}
C.~Gu, C.~Sun, D.~A. Ross, C.~Vondrick, C.~Pantofaru, Y.~Li,
  S.~Vijayanarasimhan, G.~Toderici, S.~Ricco, R.~Sukthankar, C.~Schmid, and
  J.~Malik.
\newblock Ava: A video dataset of spatio-temporally localized atomic visual
  actions.
\newblock In {\em Proceedings of the IEEE Conference on Computer Vision and
  Pattern Recognition (CVPR)}, 2018.

\bibitem{hou2017tube}
R.~Hou, C.~Chen, and M.~Shah.
\newblock Tube convolutional neural network (t-cnn) for action detection in
  videos.
\newblock In {\em Proceedings of the IEEE International Conference on Computer
  Vision (ICCV)}, 2017.

\bibitem{kay2017kinetics}
W.~Kay, J.~Carreira, K.~Simonyan, B.~Zhang, C.~Hillier, S.~Vijayanarasimhan,
  F.~Viola, T.~Green, T.~Back, P.~Natsev, M.~Suleyman, and A.~Zisserman.
\newblock The kinetics human action video dataset.
\newblock {\em CoRR}, abs/1705.06950, 2017.

\bibitem{lin2017feature}
T.-Y. Lin, P.~Dollar, R.~Girshick, K.~He, B.~Hariharan, and S.~Belongie.
\newblock Feature pyramid networks for object detection.
\newblock In {\em Proceedings of the IEEE Conference on Computer Vision and
  Pattern Recognition (CVPR)}, 2017.

\bibitem{qiu2017learning}
Z.~Qiu, T.~Yao, and T.~Mei.
\newblock Learning spatio-temporal representation with pseudo-3d residual
  networks.
\newblock In {\em Proceedings of the IEEE International Conference on Computer
  Vision (ICCV)}, 2017.

\bibitem{ren2015faster}
S.~Ren, K.~He, R.~Girshick, and J.~Sun.
\newblock Faster {R-CNN}: {T}owards real-time object detection with region
  proposal networks.
\newblock In {\em Advances in Neural Information Processing Systems (NIPS)},
  2015.

\bibitem{russakovsky2015imagenet}
O.~Russakovsky, J.~Deng, H.~Su, J.~Krause, S.~Satheesh, S.~Ma, Z.~Huang,
  A.~Karpathy, A.~Khosla, M.~Bernstein, A.~C. Berg, and L.~Fei-Fei.
\newblock {ImageNet Large Scale Visual Recognition Challenge}.
\newblock {\em International Journal of Computer Vision (IJCV)}, 2015.

\bibitem{sigurdsson2016hollywood}
G.~A. Sigurdsson, G.~Varol, X.~Wang, A.~Farhadi, I.~Laptev, and A.~Gupta.
\newblock Hollywood in homes: Crowdsourcing data collection for activity
  understanding.
\newblock In {\em Proceedings of the European Conference on Computer Vision
  (ECCV)}, 2016.

\bibitem{sun2017revisiting}
C.~Sun, A.~Shrivastava, S.~Singh, and A.~Gupta.
\newblock Revisiting unreasonable effectiveness of data in deep learning era.
\newblock In {\em Proceedings of the IEEE International Conference on Computer
  Vision (ICCV)}, 2017.

\bibitem{tran2018closer}
D.~Tran, H.~Wang, L.~Torresani, J.~Ray, Y.~LeCun, and M.~Paluri.
\newblock A closer look at spatiotemporal convolutions for action recognition.
\newblock In {\em Proceedings of the IEEE Conference on Computer Vision and
  Pattern Recognition (CVPR)}, 2018.

\bibitem{wang2016temporal}
L.~Wang, Y.~Xiong, Z.~Wang, Y.~Qiao, D.~Lin, X.~Tang, and L.~Van~Gool.
\newblock Temporal segment networks: Towards good practices for deep action
  recognition.
\newblock In {\em Proceedings of the European Conference on Computer Vision
  (ECCV)}, 2016.

\bibitem{wang2017non}
X.~Wang, R.~Girshick, A.~Gupta, and K.~He.
\newblock Non-local neural networks.
\newblock In {\em Proceedings of the IEEE Conference on Computer Vision and
  Pattern Recognition (CVPR)}, 2018.

\bibitem{yh_ava_submit}
T.~Yao and X.~Li.
\newblock Yh technologies at activitynet challenge 2018.
\newblock {\em arXiv preprint arXiv:1807.00686}, 2018.

\end{thebibliography}
}
\end{document}